\documentclass[letterpaper, 10 pt, conference]{ieeeconf}  

\IEEEoverridecommandlockouts                              

\usepackage{graphics} 
\usepackage{epsfig} 
\usepackage{mathptmx} 
\usepackage{times} 
\usepackage{amsmath} 
\usepackage{amssymb}  
\usepackage{algorithm}
\usepackage{algpseudocode}
\usepackage{subcaption}

\usepackage{tikz}
\usetikzlibrary{fit}
\tikzset{%
  highlight/.style={rectangle,rounded corners,fill=red!15,draw,
    fill opacity=0.5,thick,inner sep=0pt}
}
\newcommand{\tikzmark}[2]{\tikz[overlay,remember picture,
  baseline=(#1.base)] \node (#1) {#2};}
\newcommand{\Highlight}[1][submatrix]{%
    \tikz[overlay,remember picture]{
    \node[highlight,fit=(left.north west) (right.south east)] (#1) {};}
}

\title{\LARGE \bf
Towards a Surgeon-in-the-Loop Ophthalmic Robotic Apprentice using Reinforcement and Imitation Learning
}

\author{Amr Gomaa$^{*}$, Bilal Mahdy$^{*}$, Niko Kleer$^{*}$ and Antonio Kr{\"u}ger$^{*}$
\thanks{$^{*}$All authors are with the German Research Center for Artificial Intelligence (DFKI) and Saarland Informatics Campus, Saarland University, 66123 Saarbrucken, Germany
        {\tt\small {firstName.lastName}@dfki.de}}}

\begin{document}

\maketitle
\thispagestyle{empty}
\pagestyle{empty}

\begin{abstract}
    Robot-assisted surgical systems have demonstrated significant potential in enhancing surgical precision and minimizing human errors. However, existing systems cannot accommodate individual surgeons' unique preferences and requirements. Additionally, they primarily focus on general surgeries (e.g., laparoscopy) and are unsuitable for highly precise microsurgeries, such as ophthalmic procedures. Thus, we propose an image-guided approach for surgeon-centered autonomous agents that can adapt to the individual surgeon's skill level and preferred surgical techniques during ophthalmic cataract surgery. Our approach trains reinforcement and imitation learning agents simultaneously using curriculum learning approaches guided by image data to perform all tasks of the incision phase of cataract surgery. By integrating the surgeon's actions and preferences into the training process, our approach enables the robot to implicitly learn and adapt to the individual surgeon's unique techniques through surgeon-in-the-loop demonstrations. This results in a more intuitive and personalized surgical experience for the surgeon while ensuring consistent performance for the autonomous robotic apprentice. We define and evaluate the effectiveness of our approach in a simulated environment using our proposed metrics and highlight the trade-off between a generic agent and a surgeon-centered adapted agent. Finally, our approach has the potential to extend to other ophthalmic and microsurgical procedures, opening the door to a new generation of surgeon-in-the-loop autonomous surgical robots. We provide an open-source simulation framework for future development and reproducibility at https://github.com/amrgomaaelhady/CataractAdaptSurgRobot.
\end{abstract}

\section{Introduction}

Several decades ago, surgical robots were considered an implausible and unachievable concept; however, the da Vinci Surgical Robotic System~\cite{dimaio2011vinci} has been assisting surgeons to perform millions of operations over the past decade~\cite{intuitive2022Davinci}. Similarly, with the current advancement in artificial intelligence and machine learning, autonomous surgical robots could be a plausible, feasible, and achievable concept. In a review by Datta et al.~\cite{DATTA2021329}, it was shown that reinforcement learning (RL) has been a cornerstone for the automation of surgical robots over the past decade.
Although some existing tools bridge the gap between reinforcement learning algorithms and robotic surgery simulations, such as dVRL~\cite{dVRL}, methodologies that are concluded with positive results usually restrict the use case to a specific narrowed part of the surgical environment. In their review, Datta et al.~\cite{DATTA2021329} limit the scope of reinforcement learning scenarios to auxiliary tasks within surgical protocols, such as determining intravenous (IV) dosage or patient-assisted ventilation monitoring. However, more physically interactive applications often face limitations when the environment becomes too complex for the agent to navigate. Bourdillon et al.~\cite{rl_virtual_robotic_surgery} combined RL in a virtual robotic surgery simulation environment to train a proximal policy optimization (PPO) agent to operate a pair of scissors on a tissue-like object. The reliability of the results was positive when motion was restricted to a single axis but dropped significantly when more than one was introduced. Other applications of reinforcement learning in surgical procedures such as~\cite{nguyen_pinch,su_multicamera,barnoy_lean_rl,ning_ultrasound,baek_pathplanning,gao_gesture_recognition,xu_surrol} are also similarly bound by complexity constraints or limited function, which we are overcoming using our method of combining RL and imitation learning (IL) through curriculum learning.

\begin{figure}[t]
    \centering
    \includegraphics[width=\linewidth]{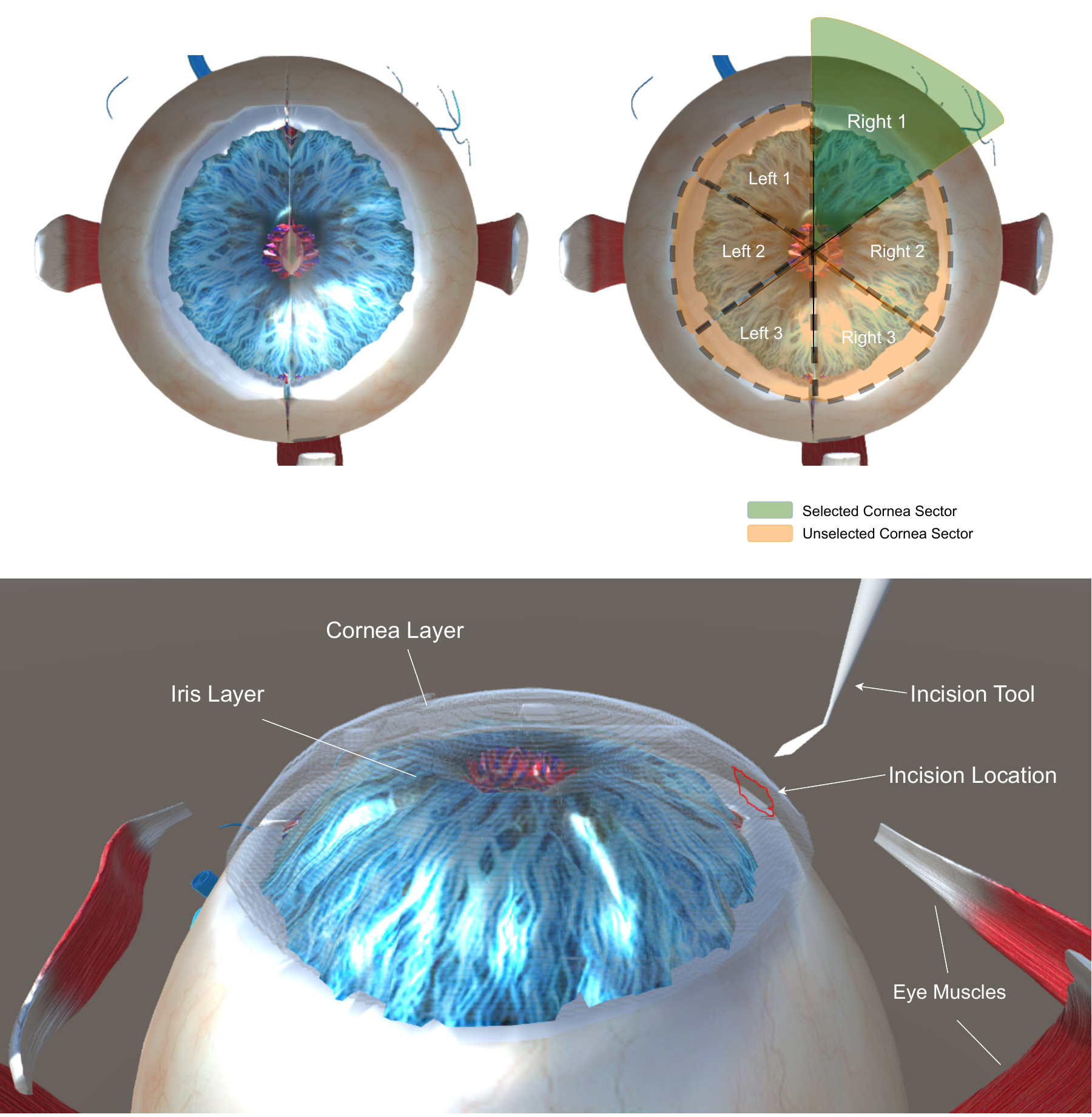}
    \caption{Our cataract surgery simulation model. Top: Cornea illustrative sectorization for personalized approach modeling. Bottom: The surgical tool performing an incision in the cornea (highlighted in red).}
    \label{fig:teaser}
\end{figure}

In addition to enhancing surgical precision and patient safety, there is growing interest in the integration of advanced technologies, such as surgical robots, into the field of cataract surgery~\cite{Pandey2019-rb} (see Fig.~\ref{fig:teaser}). The potential of combining imitation learning with robotic assistance holds great promise for revolutionizing the surgical landscape. Although in this work we primarily consider improving surgical techniques through imitation learning in a simulated environment, we also highlight the future prospect of transferring the acquired knowledge and skills to a physical surgical robot. In other words, by pretraining an imitation learning agent to perform cataract surgery with increased precision within a realistic simulation environment, we provide a base model for the seamless transfer of these skills to a physical robotic surgical system through Transfer Learning~\cite{torrey2010transfer} and Curriculum Learning~\cite{bengio2009curriculum} techniques. This integration of robotic assistance into surgery can further amplify surgical precision, minimize human error, and allow highly controlled and precise surgical maneuvers. By exploring the correlation and trade-off between imitation learning, virtual simulation, and robotic assistance, our research aims to improve the outcome of cataract surgery by improving surgical precision and reducing surgeon errors through the utilization of autonomous and semi-autonomous surgical robots.

In summary, our \textbf{contributions} are threefold, as follows. 
\begin{enumerate}
    \item We propose a novel approach for customizing the surgical behaviour of an imitation learning agent based on visual input and recorded demonstrations, tailoring it to individual surgeon techniques through \textit{Transfer Learning}~\cite{torrey2010transfer} and \textit{Curriculum Learning} techniques~\cite{bengio2009curriculum}.
    \item We introduce a comprehensive 3D simulation framework in \textit{Unity}, encompassing a precise and appropriately scaled model of the human eye, a set of scaled microsurgery tools, and a cohesive implementation of incision/collision physics that govern interactions between the tools and different eye components.
    \item We demonstrate and evaluate an agent execution of the incision phase of the ophthalmic cataract surgery and adaptability to surgeon specific style.
\end{enumerate} 

\section{Background and Related Work}
\label{sec:background}	

    \subsection{Surgical Microscopy and Robotics}

    Surgery, as a medical specialty, uses operative manual and instrumental techniques applied to the human body to treat a patient’s injury or pathological condition. Such microscopic procedures are performed manually by experienced surgeons with the help of a surgical microscope and require, in addition to trained hand-eye coordination and the planned procedure, machines and technology to make time-critical and risky interventions feasible.
    More recently, with the introduction of surgical robotic systems, the operating surgeon manually configures surgical robotic imaging microscopes to ensure optimal support through the imaging properties of the visual device directed at the surgical zone. However, imaging requirements in terms of positioning, viewing, and focus plane may change throughout the operation, making it necessary for surgeons to adapt the configuration of the surgical microscope. Since adjustments must be performed manually, a short-term interruption of the operation becomes necessary. Individual risk considerations are made in each case as to whether visual section correction is effective in the present surgical situation, which could negatively affect the outcome of the surgery if taken incorrectly.
    Hence, there have been recent advances in the hand-free configuration of the microscope, such as mouth-controlled joysticks or foot control with the help of pedals~\cite{holly1976mouth,sindou2009practical,afkari2014potentials}. However, these approaches are impractical and allow too few degrees of freedom to meet the complex requirements for the various settings of the microscopic frame~\cite{gomaa2023teach}, unlike a fully automated approach as presented in this work.
    
    \subsection{Reinforcement Learning (RL)}

    RL dates back to 1983 when it was first introduced in the research areas of cybernetics, statistics, psychology, and neuroscience~\cite{kaelbling1996reinforcement,kiumarsi2017optimal}. In 1996, Kaeblings et al.~\cite{kaelbling1996reinforcement} highlighted the use of RL in artificial intelligence (AI) research as a way of teaching an agent how to complete a task based only on reward and punishment (i.e., without specifying how the task is actually achieved). With recent advances in computational power, researchers were able to combine RL with deep learning approaches and apply these algorithms to specific tasks with a predefined reward system (e.g., Atari games~\cite{mnih2013playing}). 
    There are different types of reinforcement learning algorithms commonly used, each with its own characteristics and applications. One type is value-based methods, where the agent learns to estimate the value of different states or state-action pairs. These algorithms, such as Q-learning~\cite{jang2019qlearning} and deep Q-Networks (DQN)~\cite{mnih2013playing}, aim to find an optimal value function that guides the agent's decision-making. Another type is policy-based methods, which directly learn a policy, a mapping from states to actions. These algorithms, including REINFORCE~\cite{Williams1992reinforce} and Proximal Policy Optimization (PPO)~\cite{schulman2017proximal}, explore the policy space and optimize the agent's policy to maximize the expected cumulative rewards.
    However, due to the computationally expensive nature of RL, researchers were unable to apply such algorithms to complex tasks. Additionally, current advances in the RL approach depend on hand-crafted sensitive, unstable reward systems that require high computational power and extended training time. Therefore, there is a need for a reward-free learning algorithm, such as the combination of imitation learning and reinforcement learning used in this work.

    \subsection{Imitation Learning}
    
    One of the methodologies used to solve the dilemma of defining the reward function with RL is imitation learning (IL) or learning from demonstrations, focusing on training agents to mimic expert behavior. Unlike RL, where agents learn through trial and error, IL uses expert demonstrations as guidance. In this approach, an agent learns a policy by observing and mimicking the actions performed by human or expert demonstrators. By learning from preexisting knowledge and demonstrations, imitation learning offers several advantages, such as faster convergence and improved sample efficiency compared to pure reinforcement learning~\cite{zheng2022imitation,DBLP:conf/hitlaml/GomaaM23}. It finds applications in various domains, including robotics~\cite{fang2019survey, lopes2007affordance, ratliff2007imitation, zhu2018reinforcement, pore2021learning}, autonomous driving~\cite{codevilla2018end, pan2017agile, zhang2016query}, and virtual simulation~\cite{bhattacharyya2018multi}. Imitation learning algorithms, such as Behavioral Cloning~\cite{torabi2018behavioral}, Inverse Reinforcement Learning (IRL)~\cite{arora2020survey}, and Generative Adversarial Imitation Learning (GAIL)~\cite{ho2016generative}, provide frameworks for effectively transferring expert knowledge to autonomous agents. However, imitation learning also faces challenges such as handling distribution shifts between expert demonstrations and agent exploration, generalization to novel situations, and addressing the trade-off between imitation accuracy and exploration for robust decision making~\cite{zheng2022imitation}. Therefore, our work hierarchically combines imitation and reinforcement learning to achieve a trade-off between the two methodologies within adequate performance levels.

    \subsection{Ophthalmic Cataract Surgery}
    Cataract surgery is a standard procedure to treat cataracts, a condition characterized by clouding the eye's natural lens~\cite{davis2016evolution}. The clouded lens is removed and replaced with an artificial intraocular lens (IOL) during surgery. The procedure is generally performed under local anesthesia and, in some cases, with mild sedation~\cite{nouvellon2010anaesthesia}. Surgeons make a small incision in the cornea, the transparent outer layer of the eye, and access the clouded lens. The surgeon breaks down and removes cloudy lens fragments through various techniques, such as phacoemulsification~\cite{buratto2003phacoemulsification}. Once the lens is completely removed, an IOL is inserted into the eye to restore clear vision. Cataract surgery is known for its high success rate and relatively quick recovery period, allowing individuals to regain improved vision and resume daily activities with minimal disruption. Therefore, the systematic procedure and high precision of cataract surgery make it a suitable benchmark for a semi- or fully-automated robotic surgical apprentice.

\section{Methodology}
\label{sec:methodology}

	The methodology presented in this paper outlines an innovative approach to train an imitation learning agent capable of performing the cataract surgery incision task within a customized Unity simulation environment. To augment surgical training and improve surgical outcomes, we propose a comprehensive framework that uses state-of-the-art techniques in reinforcement learning and imitation learning. By combining expert demonstrations, policy optimization algorithms, and a realistic surgical simulator, our methodology empowers the agent to learn complex surgical maneuvers, decision-making processes, and delicate interactions with surgical tools. This section details the key components of our approach, including data collection, pre-processing, training pipeline, simulation environment design, and performance evaluation metrics.

    \begin{figure}[t]
        \centering
        \includegraphics[width=0.9\linewidth]{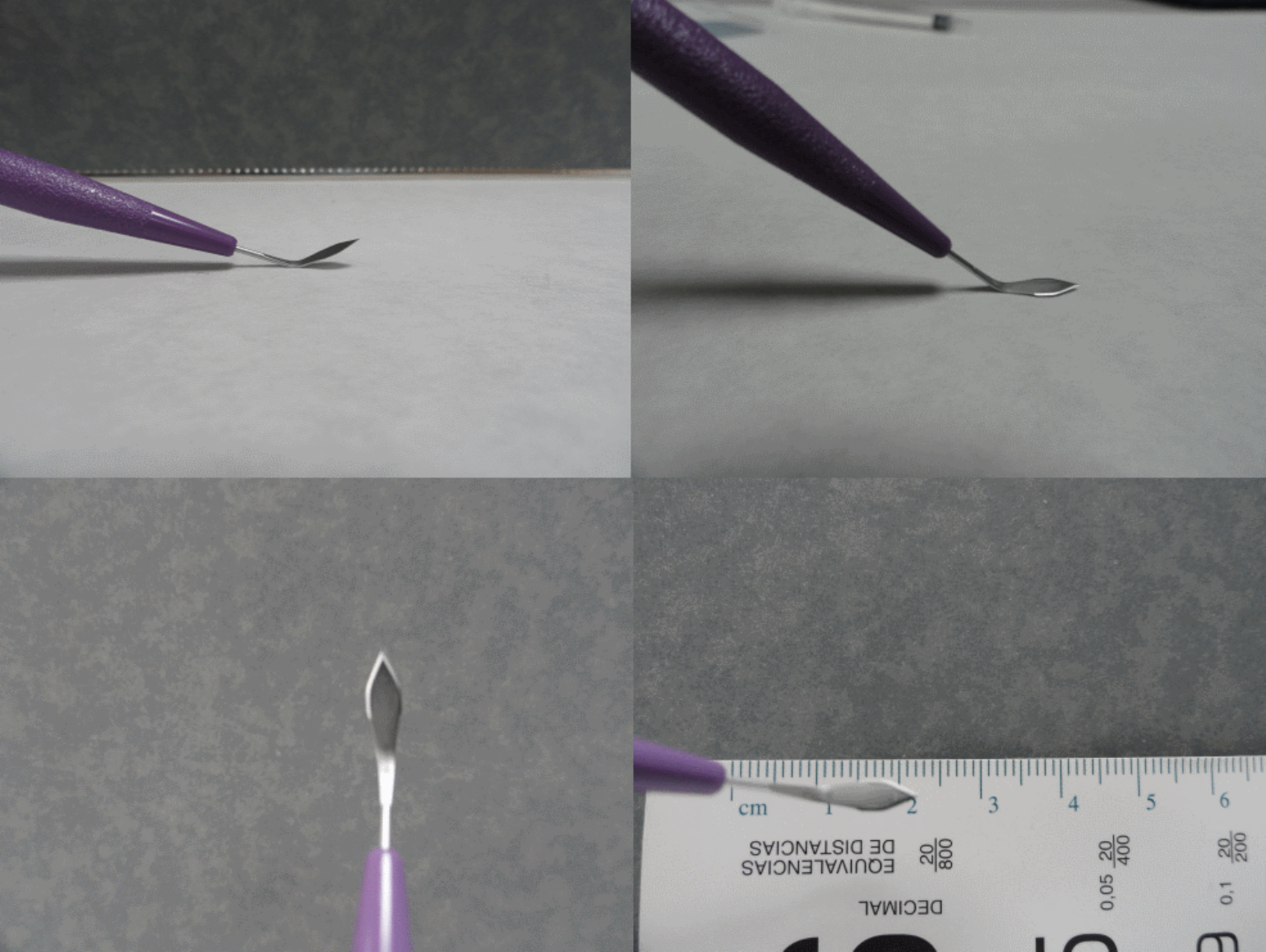}
        \caption{A 2.75mm Keratome Ophthalmic Knife mapped and digitalized in 3D by~\cite{coca2013models}.}
        \label{fig:tool_model}
    \end{figure}

\begin{figure}[b]
    \centering
    \includegraphics[width=0.8\linewidth]{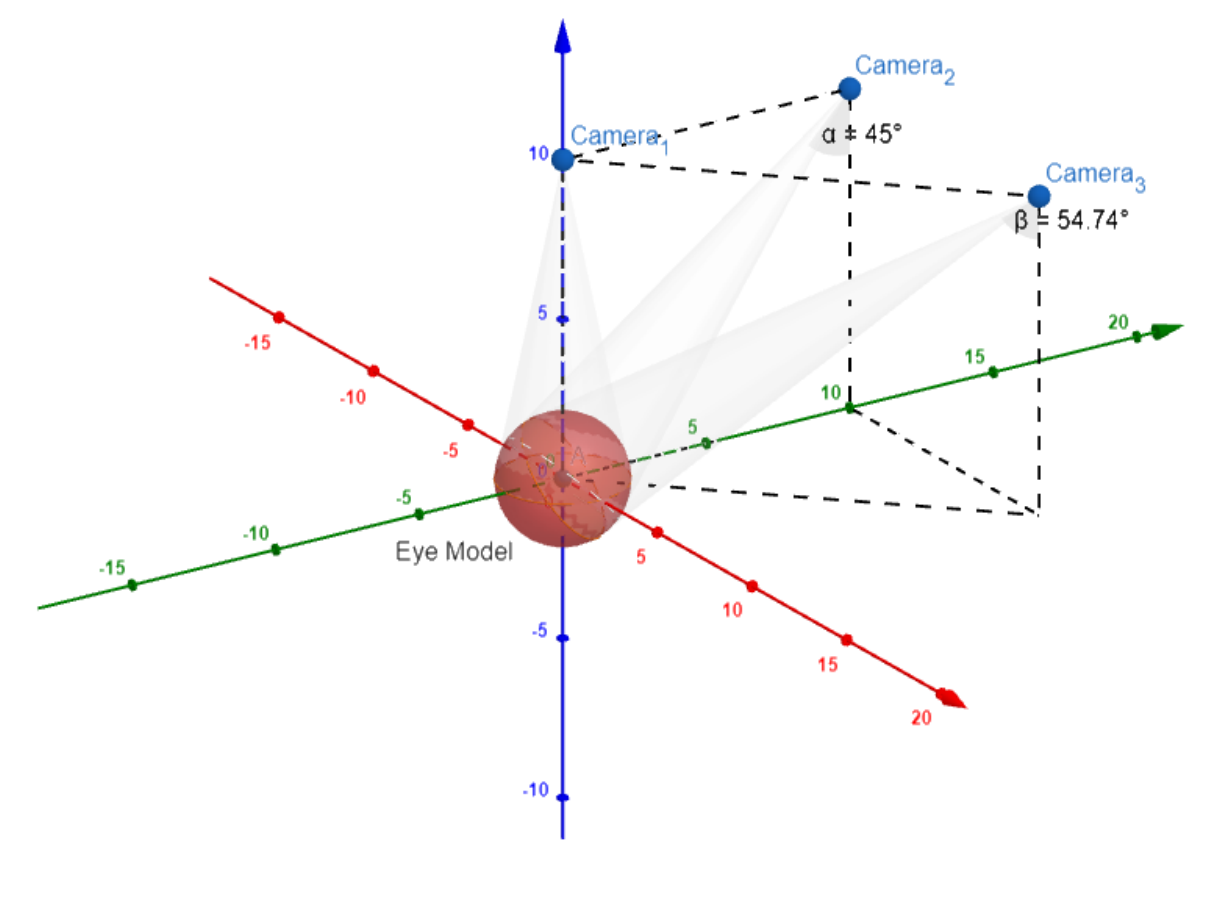}
    \caption{Camera setup for comprehensive surgical scene perception. The top-view camera (Camera 1) provides an overhead vantage point, capturing the surgical field and the overall context. The upper-side view camera (Camera 2) offers a lateral perspective, enabling observation of side-specific details and depth perception. Lastly, the camera with a view in the upper corner (Camera 3) provides an angled viewpoint, enhancing spatial awareness and facilitating precise tool manipulation.}
    \label{fig:cam_positions}
\end{figure}

 \subsection{Simulation Environment}
 \label{subsec:method_simulation}
	The simulation environment utilized in this study serves as a vital component of our training methodology, providing a realistic and immersive platform for the imitation learning agent to develop surgical skills. Specifically designed for cataract surgery, the simulation environment incorporates meticulously modeled surgical tools (provided by~\cite{coca2013models}, see Fig.~\ref{fig:tool_model}), a highly accurate human eye model (see Fig.~\ref{fig:teaser}), and a three point-of-view camera system (see Fig.~\ref{fig:cam_positions}). The primary incision tool, a 2.75mm Keratome Ophthalmic Knife (shown in Fig.~\ref{fig:tool_model}), is intricately modeled to capture the physical properties and functionalities of the real tool. 
    
     The eye model used in this study comprises a comprehensive representation of its various anatomical components, including the cornea, iris, veins, arteries, and optic nerves. Each component is modeled to capture its unique characteristics and structural intricacies. The cornea, which acts as the transparent outermost layer, is reconstructed to emulate its curved shape and refractive properties. The iris, responsible for regulating the amount of light entering the eye, is depicted with its intricate patterns and radial muscle fibers. Voxelization techniques were employed to discretize the eye model into a three-dimensional grid, with each voxel measuring 0.01mm in size. This high resolution voxelization enables precise simulations of surgical procedures, such as cutting or incisions, and ensures accurate modeling of tissue interactions. The veins, arteries, and optic nerves, crucial for the eye's vascular supply and visual information transmission, are represented to mimic their branching networks and connectivity within the eye. Moreover, deformation in the eye model through cutting and incision is simulated by removing the affected voxels to show the layers behind it (e.g., blood vessels) which mimics tissue displacement in real-world physics as seen in~\cite{nair2021effectiveness}.   Therefore, our eye model provides a realistic and fine-grained simulation environment, enabling the agent to simulate and learn from intricate surgical maneuvers within the eye with a high degree of fidelity.

\subsection{Curriculum Learning Agent}
 \label{subsec:method_architecture}

    Our approach combines reinforcement learning (i.e., PPO) with imitation learning (i.e., GAIL) using a curriculum learning~\cite{bengio2009curriculum} approach. While other reinforcement methodologies could be used, PPO provides more stability to the learning process that balances the unstable nature of adversarial neural networks in GAIL~\cite{DBLP:conf/hitlaml/GomaaM23}.
    To achieve this, we use two different environment structures with increasing difficulty (i.e., low versus high number of polygons) for curriculum learning~\cite{bengio2009curriculum, narvekar2016source} as follows.
    
    \begin{figure}[t]
    \centering
    \begin{subfigure}{0.49\linewidth}
         \centering
         \includegraphics[width=\textwidth]{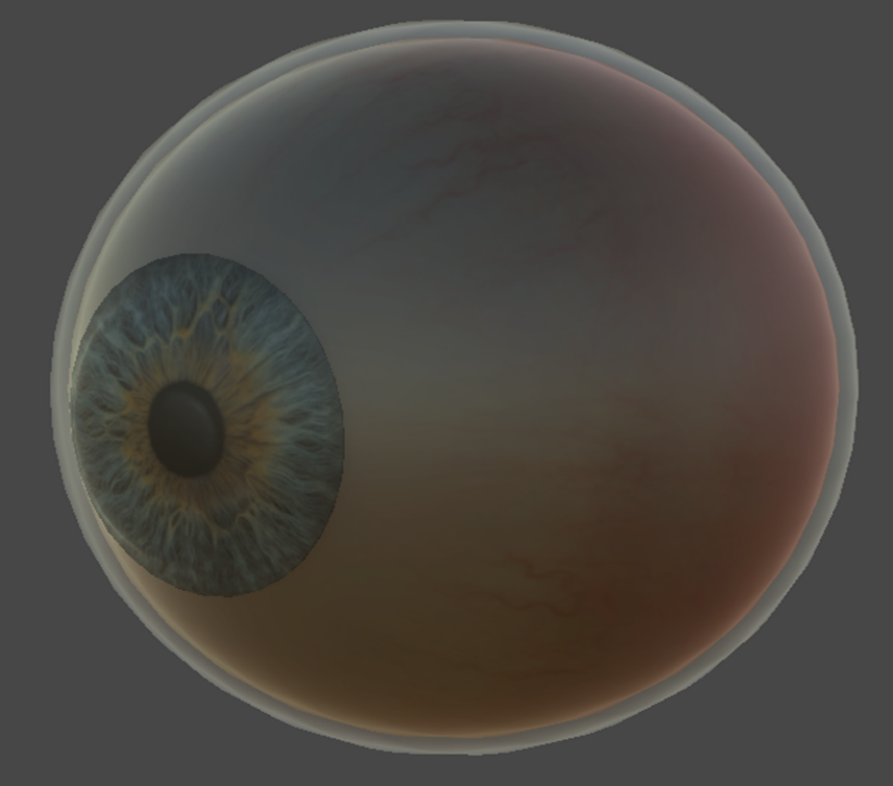}
     \end{subfigure}
     \begin{subfigure}{0.49\linewidth}
         \centering
         \includegraphics[width=\textwidth]{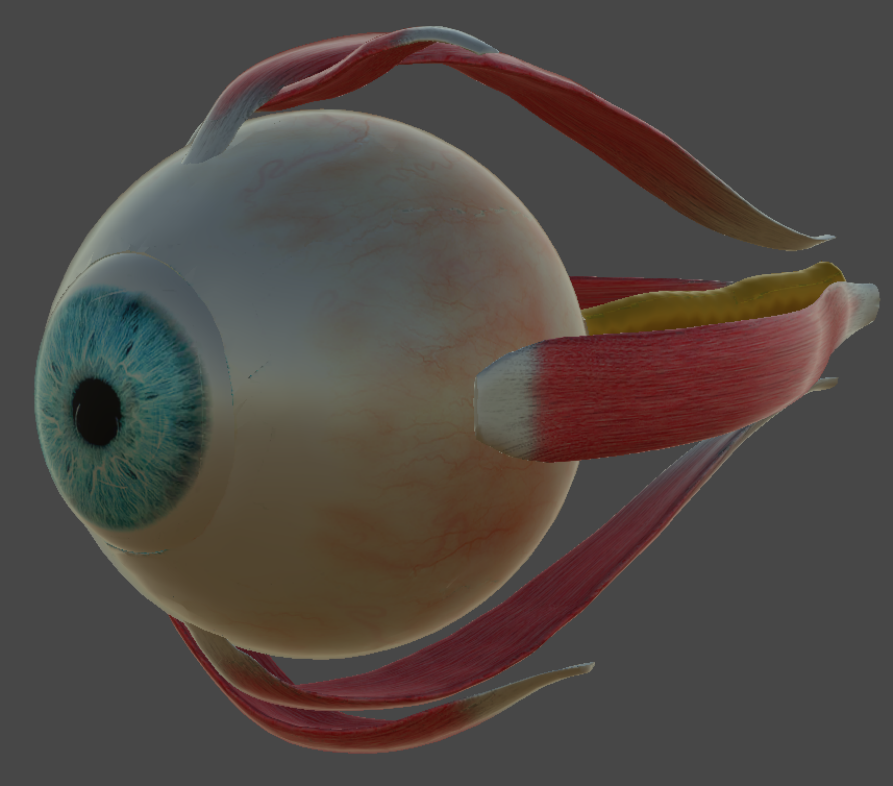}
     \end{subfigure}
    \caption{\textit{Left}: Low Poly eye model, \textit{Right}: High Poly eye model}
    \label{fig:lowpoly_vs_highpoly}
\end{figure}

\begin{figure}[b]
\centering
    \includegraphics[width=1.0\linewidth]{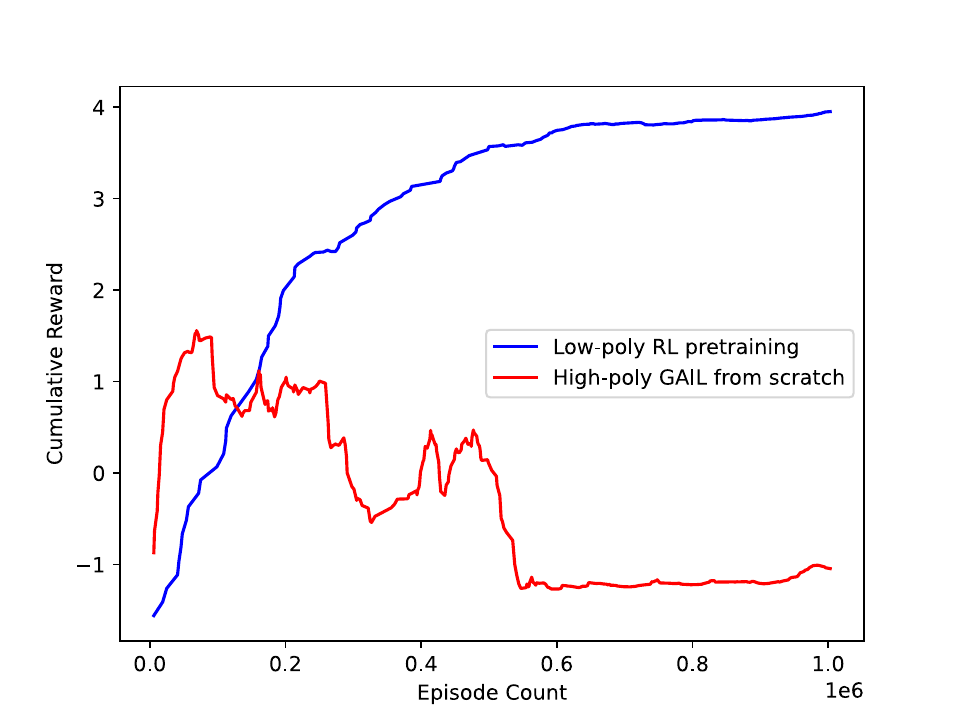}
    \caption{Cumulative reward convergence using RL pretraining on a low-poly eye model versus using a GAIL agent directly on a high-poly eye model.}
    \label{fig:gail_vs_rl}
\end{figure}

    \subsubsection{Low Poly -- Simple Learning Task}
        Our agent performs preliminary training in a significantly less complex environment, including an eye model with a low polygon count, no textures, lower degrees of freedom in tool manipulation and a more lenient reward function (see Fig.~\ref{fig:lowpoly_vs_highpoly} \textit{Left}). This acts as a stepping stone to facilitate convergence in the high poly environment as in previous work on curriculum learning. The reward function is designed so that the agent is simply rewarded for performing an incision anywhere on the cornea layer without conforming to a particular surgery technique and without brushing, cutting, or damaging another part of the eye anatomy in any way possible. This is the initial \textit{Simple Learning Task}.
        Skipping this pre-training results in the agent being unable to learn the simulation environment's complexity. As seen in Fig.~\ref{fig:gail_vs_rl}, the agent trained using traditional RL in the low poly environment can converge steadily and consistently. In contrast, the agent trained directly in the high poly environment cannot capture the visual complexity throughout the training process.

    \subsubsection{High Poly -- Complex Learning Task}
\begin{figure}[t]
    \centering
    \includegraphics[width=0.9\linewidth]{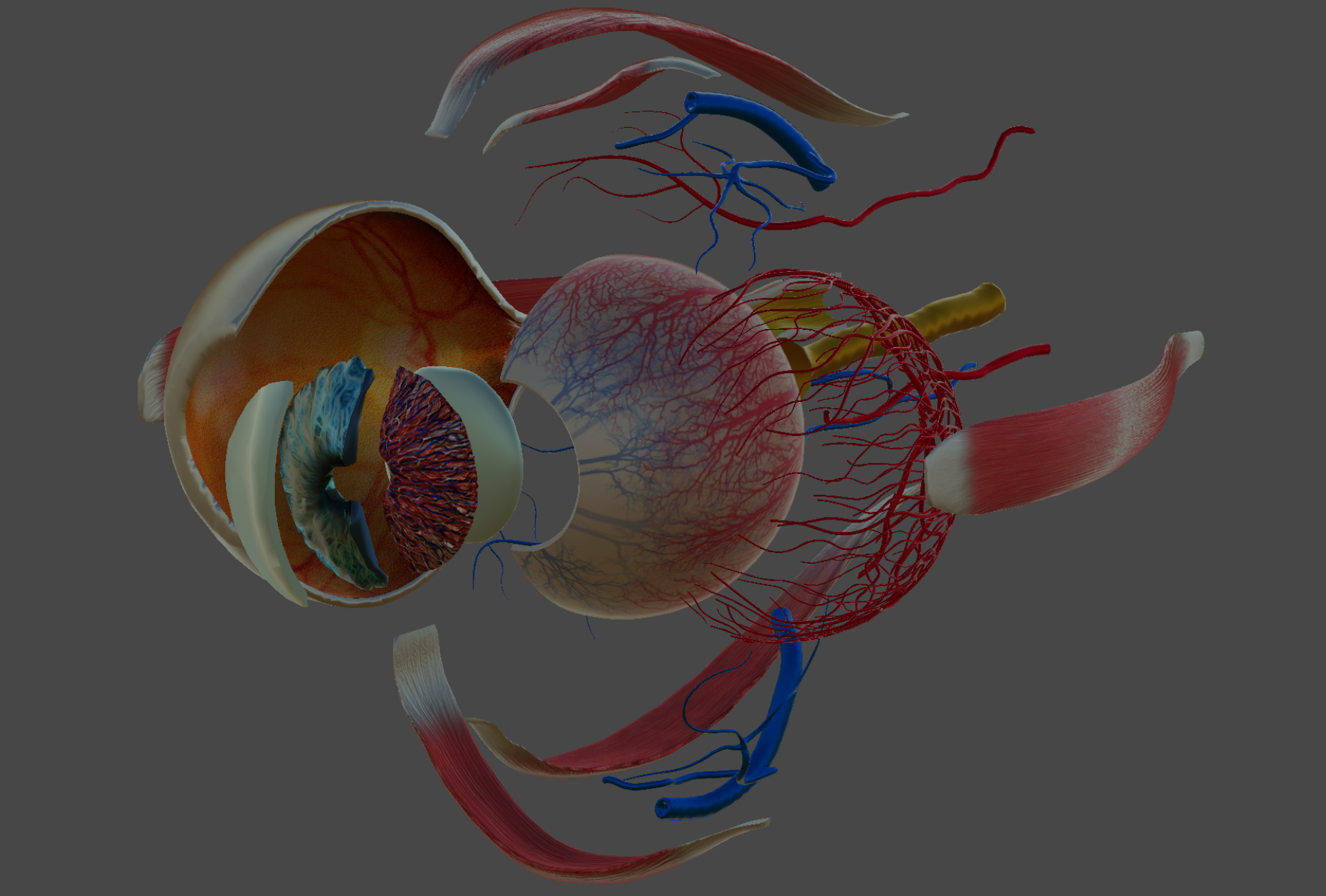}
    \caption{Internal modelled features of the eye in the High Poly environment}
    \label{fig:eye_internal}
\end{figure}

        After the agent converges in the low poly environment, it is transferred to the higher poly environment and tuned to learn its new rules. The high poly environment rewards the agent using the same conditions as the low poly environment. However, it contributes two additions as follows. A highly textured eye model with high polygon counts, and a punishment for the agent if it performs an incision that does not conform to proper cataract surgery techniques or could cause complications for the patient~\cite{chan2010complications}. A comparison of the low and high poly models is shown in Fig.~\ref{fig:lowpoly_vs_highpoly}. Note that the polygon count is not apparent in the rendered images. Additionally, Fig.~\ref{fig:eye_internal} shows the internal modelled features within the high poly model in which every component interacts dynamically with the incision tool controlled by the agent. 
        The additional punishment is presented later in the reward function of Algorithm~\ref{alg:algorithm}. This is the final \textit{Complex Learning Task}.

 \subsection{Learning Environment}
 The learning environment is built within the Unity simulation in Sec.\ref{subsec:method_simulation}, in which the task $\epsilon = \langle S, O, A, p, r, \gamma \rangle $ is represented as a Partially Observable Markov Decision Process (POMDP)~\cite{murphy2000survey, sutton1998introduction}.
\begin{itemize}
  \item \textbf{State space} $S$ contains the ground-truth state of the environment, is defined and operated by the Unity simulation and remains hidden to the agent. At each timestep $t$, the environment's state $s_t \in S$ yields a set of observations $o_t \in O$.
  \item \textbf{Observation space} $O$ contains the visible information given to the agent. It includes the visual observations from three cameras strategically positioned for optimal visual coverage. The position and number of cameras were optimized using a grid search approach over a nine-camera topology in various positions. The three cameras capture a top view, an upper side view, and an upper corner view, providing a comprehensive perspective of the surgical scene (shown in Fig.~\ref{fig:cam_positions}). These observations are obtained as colored pixel data with a resolution of $128\times128\times3$, allowing the agent to perceive fine details and spatial relationships between different environmental elements. Additionally, the exact position of the incision tool is included in every observation timestep $o_t$, enabling it to have real-time awareness of the tool's location within the surgery simulation. Finally, $o_t$ also includes the Euclidean distance between the tool's edge and the eye model for proximity and contact calculation.
  \item \textbf{Action space} $A$ is represented as the 3D affine transformation matrix $M$ (shown in Equation~\ref{equ:transformation_matrix}), which rotates and translates the incision tool within the simulation. 
 \begin{equation}
     M =
      \begin{Bmatrix}
    \tikzmark{left}{$a_{11}$} & a_{12} & a_{13} & a_{14}\\
    a_{21} & a_{22} & a_{23} & a_{24}\\
    a_{31} & a_{32} & \tikzmark{right}{$a_{33}$} & a_{34}\\
    0 & 0 & 0 & 1
    \end{Bmatrix}
    \Highlight[first]
    \label{equ:transformation_matrix}
 \end{equation}

 where the upper-left $3 \times 3$ highlighted sub-matrix represents a rotation transform (i.e., orientation). The last column of the matrix ($a_{14}$, $a_{24}$, and $a_{34}$) represents a translation transform (i.e., position). The transformation action $a_t \in A$ is projected onto the environment by a matrix multiplication \textbf{$p_{T_{t}} = M p_{T_{t-1}}$}.

    \item \textbf{Transition Behaviour} $p(s_{t+1} | s_t, a_t)$ defines how the state $s_t \in S$ changes every timestep after the agent performs action $a_t \in A$. Since the agent is designed in a model-free approach, this is controlled by the Unity engine, and not explicitly learned by the agent.
   
   \item \textbf{Environment Reward} $r$ represents the reward function given to the agent at every timestep, shown in Equation~\ref{equ:reward_function}.

  \begin{equation}
        r_{env_t}= 
            \begin{cases}
                R_{Correct},& \text{if hit correct \& } hit_{c} < \beta, \\
                R_{Success},& \text{if hit correct \& } hit_{c} = \beta, \\
                R_{Fail},& \text{if hit wrong},\\
                -k_{time} + \Delta{d_t},& \text{otherwise}
            \end{cases}
        \label{equ:reward_function}
    \end{equation}

    where $k_{time}$ is a constant time delay penalty given to the agent, $\Delta{d_t}$ represents the change in the euclidean distance between the tool position $p_{T}$ and the geometric center of the cornea $p_{C}$, calculated as $\Delta{d_t} = ||p_{T_{t-1}} - p_{C}|| - ||p_{T_{t}} - p_{C}||$, $R_{Correct}$, $R_{Success}$, and $R_{Fail}$ are the reward values for cutting correct tissue, completing the incision, and failing the process by cutting wrong tissue, respectively. All rewards, penalties, and hyperparameters were fine-tuned using a stochastic optimization approach (optimized values are available in our framework).

    \item \textbf{Discount Factor} $\gamma$ is the factor by which future rewards are discounted each timestep.
\end{itemize}

 After training the agent on the Low Poly environment, We employ curriculum learning methods to train the agent as a foundational benchmark for training various agents, each with distinct reward ratios comprising a blend of GAIL and standard RL rewards. These reward ratios are denoted as strength factors $\lambda_{gail}$ and $\lambda_{env}$, which are used to scale the rewards generated by their respective algorithms. The total reward for each agent is subsequently calculated as the weighted sum of these individual rewards. Algorithm~\ref{alg:algorithm} shows the fine-tuning procedure for the adapted agents. Table~\ref{tab:ratios} highlights the specific ratios of $\lambda_{gail}$:$\lambda_{env}$ applied to the fine-tuned agents.

\begin{algorithm}[t]
    \caption{Cataract Surgery Adapted GAIL Agent}
    \label{alg:algorithm}
    \begin{algorithmic}
       \Require Expert Observation-Action pairs $(O_E,A_E)$, Pre-trained policy $\pi_o$, Reward function $env(o,a)$, GAIL discriminator $D(o,a)$, GAIL strength factor $\lambda_{gail}$, Environment reward strength factor $\lambda_{env}$
       \Ensure Optimized agent policy $\pi^*$
        \While {true}
            \State Sample action $a_t$ from policy $\pi_0$ and observation $o_t$
            \State \texttt{$r_{env_t} = env(o_t,a_t)$}  \Comment{Environment Reward (Eq.\ref{equ:reward_function})}
            \State \texttt{$r_{gail_t} = D(o_t,a_t|O_E,A_E)$}  \Comment{GAIL Reward}
            \State \texttt{$r_{total_t} = r_{gail_t}\lambda_{gail} + r_{env_t}\lambda_{env}$}
            \State Use PPO to optimize agent $\pi^*$ using reward $r_{total_t}$
            \If{$r_{env_t} == R_{Success}$ OR $r_{env_t} == R_{Fail}$}
                \State \textbf{break}
            \EndIf
        \EndWhile
        \State \textbf{return}  $\pi^*$
    \end{algorithmic}
\end{algorithm}

 \begin{table}[t]
 \caption{Strength factors ($\lambda$) for GAIL and RL rewards for each fine-tuned agent.}
     \centering
     \resizebox{0.7\linewidth}{!}{
     \begin{tabular}{c|c|c}
          \hline
          \textbf{Agent Type} & $\lambda_{gail}$ & $\lambda_{env}$ \\
          \hline \hline
          NonAdaptAgent & 0.0 & 1.0 \\
          BalancedAdaptAgent & 0.5 & 0.5 \\
          HighAdaptAgent & 0.7 & 0.3 \\
          PurelyAdaptAgent & 1.0 & 0.0 \\
          \hline
     \end{tabular}
     }
     \label{tab:ratios}
 \end{table}

\begin{figure*}[t]
         \centering
         \includegraphics[width=0.8\textwidth]{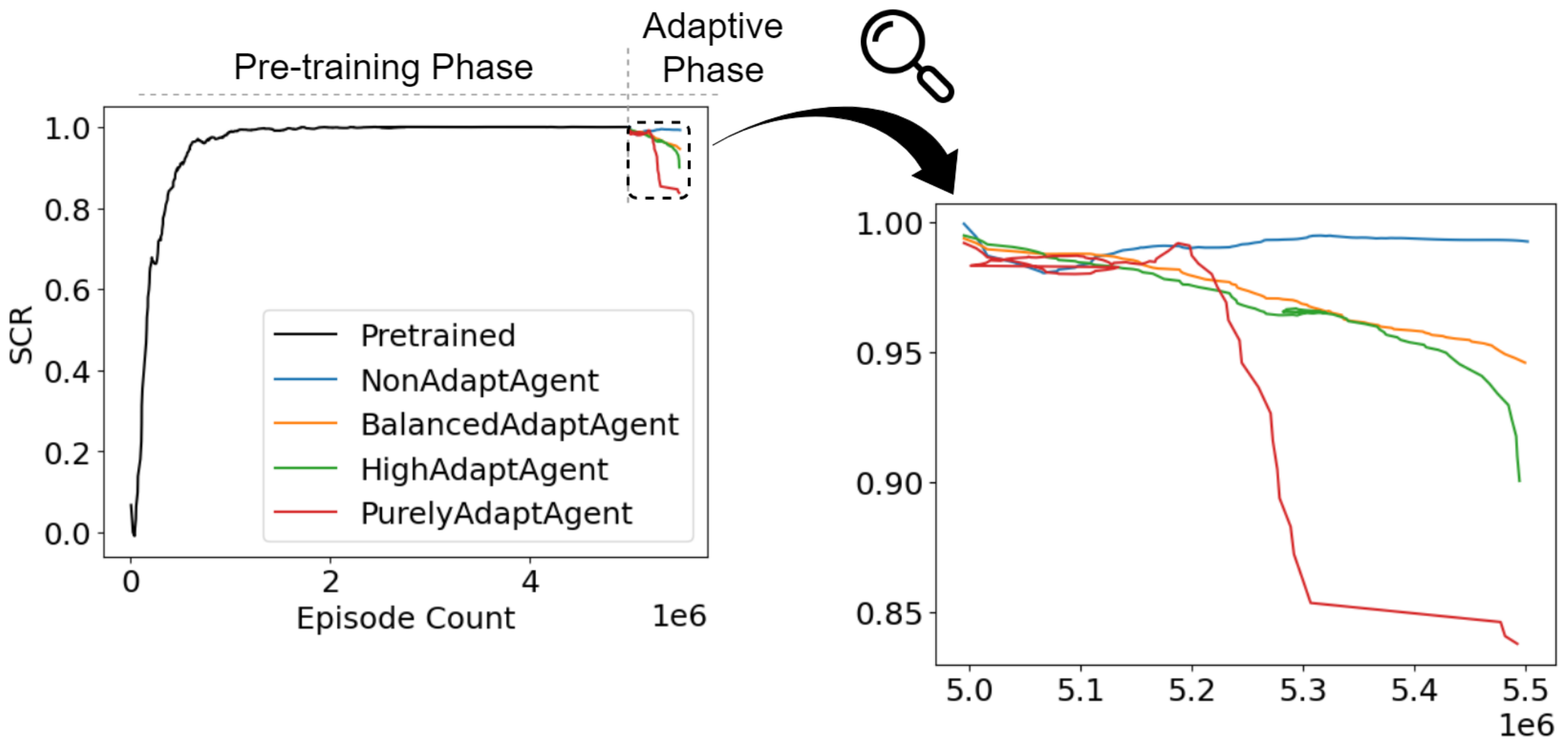}
    \caption{\textit{Surgery Completion Rate} (\textbf{SCR}) results for the complete training cycle of four agents with different strength factor ratios $\lambda_{gail}$ and $\lambda_{env}$ (see Table~\ref{tab:ratios}).}
    \label{fig:pretraining}
\end{figure*}

\section{Results and Discussion}
\label{sec:results}

	The experimental setup aims to develop a personalized model that could adapt to the individual approach of a surgeon during cataract surgery. To emulate this personalized adaptation, the cornea was divided into sectors, representing distinct regions of the surgical site. Expert demonstrations were recorded for each sector, capturing individual surgeons' specific techniques and approaches. By training the imitation learning agent on these sector-specific demonstrations, our motivation is to create a model that could learn and emulate an individual surgeon's unique approach and expertise. This personalized approach aims to improve surgical precision by allowing the agent to adapt its policy to match the surgeon's preferences and techniques in each sector of the cornea.

 \subsection{Performance Metrics}

 For this experiment, we introduce two principal metrics for measuring performance: \textit{Surgery Completion Rate} (\textbf{SCR}) and \textit{Adaptive Surgery Success Rate} (\textbf{AdSSR}). The former defines the rate at which the agent can complete an incision without damaging any part of the eye model while also conforming to proper cataract surgery techniques or could present complications to the patient~\cite{chan2010complications}. The latter defines the rate at which the agent can follow the correct technique demonstrated by the expert surgeon when performing the incision.

\begin{table*}[t]
 \caption{AdSSR values for adaptive agents fine-tuned on surgeon demonstrations. Values correspond to the mean rate and standard error (\textit{AdSSR\_mean}~$\pm$~\textit{std\_err}) at which the agent can follow their designated surgeon's approach.}
     \centering
     \resizebox{\linewidth}{!}{
     \begin{tabular}{c|c|c|c|c}
          \hline
          \textbf{Surgeon Adaptation} & \textbf{NonAdaptAgent} & \textbf{BalancedAdaptAgent} & \textbf{HighAdaptAgent} & \textbf{PurelyAdaptAgent}  \\
          \hline \hline
          Left Eye Sector (Any) & 0.28~$\pm$~0.002 & 0.28~$\pm$~0.002 & 0.44~$\pm$~0.007 & 0.70~$\pm$~0.012  \\
          \hline 
          Left 1 (Upper) & 0.07~$\pm$~0.019 & 0.07~$\pm$~0.019 & 0.11~$\pm$~0.062 & 0.18~$\pm$~0.099   \\
          Left 2 (Middle) & 0.11~$\pm$~0.014 & 0.12~$\pm$~0.014 & 0.18~$\pm$~0.040 & 0.28~$\pm$~0.070   \\
          Left 3 (Lower) & 0.10~$\pm$~0.015 & 0.10~$\pm$~0.015 & 0.15~$\pm$~0.050 & 0.24~$\pm$~0.080    \\
          \hline \hline
          Right Eye Sector (Any) & 0.72~$\pm$~0.001 & 0.72~$\pm$~0.001 & 0.80~$\pm$~0.010 & 0.84~$\pm$~0.011   \\
          \hline
          Right 1 (Upper) & 0.19~$\pm$~0.042 & 0.19~$\pm$~0.042 & 0.21~$\pm$~0.170 & 0.22~$\pm$~0.200   \\
          Right 2 (Middle) & 0.30~$\pm$~0.038 & 0.30~$\pm$~0.034 & 0.33~$\pm$~0.160 & 0.35~$\pm$~0.180   \\
          Right 3 (Lower) & 0.23~$\pm$~0.040 & 0.24~$\pm$~0.034 & 0.26~$\pm$~0.170 & 0.27~$\pm$~0.190   \\
          \hline
     \end{tabular}
     }
     
     \label{tab:adssr}
 \end{table*}

 \begin{figure*}
    \centering
    \includegraphics[width=\textwidth]{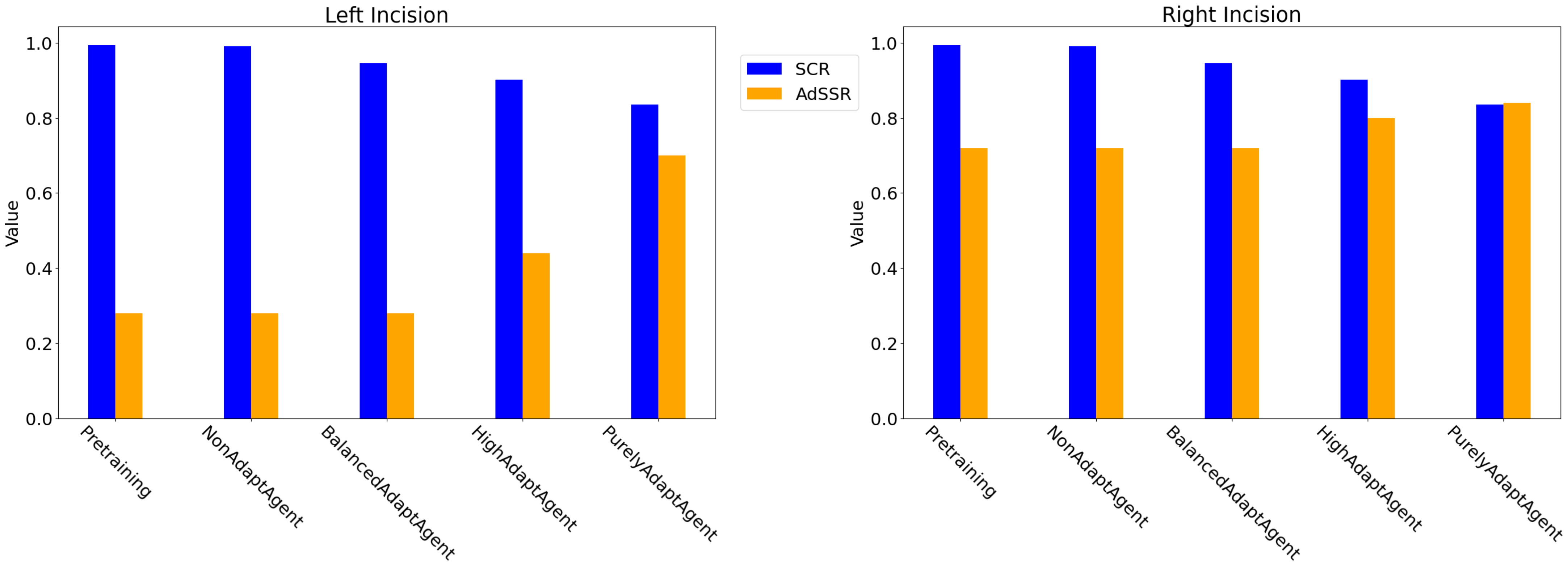}
    \caption{Comaprison between SCR and AdSSR metrics for left and right incision across all agents.}
    \label{fig:scr_vs_adssr}
\end{figure*}

 \subsection{Simulated Scenarios}

 To evaluate the adaptive agent's capacity to emulate the surgical style of a chosen surgeon, we conducted experiments across varying levels of sector resolution. In the lower sector resolution tests, we assessed the agent's capability to execute incisions on either the left or right half of the eye (i.e., two sector resolution), guided by an expert surgeon demonstrating these respective incisions. On the contrary, the higher sector resolution experiment further divides each half into three distinct sectors, resulting in a total of six sectors denoted as \textit{Left 1 to 3} and \textit{Right 1 to 3} as shown in Fig.~\ref{fig:teaser}). Note that the sectors are not uniform due to asymmetry in the eye model. Thus, the left and right sectors are not precisely halves, but they have a ratio of one to three, i.e., random chance levels for hitting the left and right sectors are 25\% and 75\%, respectively. While this is a limitation of the eye model's mesh design, it does not affect the agent performance as these sectors are neither part of the observation space $O$ nor the reward functions (i.e., $r_{env}$ and $r_{gail}$).

 \subsubsection{From Pre-training to Adapting}

 As discussed in section~\ref{subsec:method_architecture}, a pretraining operation was implemented before adapting each agent to the approach of a particular surgeon. The pretraining phase encompassed five million episodes, which was subsequently succeeded by an adaptation phase consisting of five hundred thousand episodes. Fig.~\ref{fig:pretraining} highlights the behavior of each agent during the adaptation phase for different strength factors shown in Table~\ref{tab:ratios}. It is shown that the SCR of an agent consistently decreases with the higher ratio of $\lambda_{gail}$ to $\lambda_{env}$ ending at a value of $0.835$ with the \textit{PurelyAdaptAgent} starting from the value of $0.994$ for the \textit{NonAdaptAgent}. This shows that the more the agent tries to mimic the expert surgeon, the less likely it is to successfully complete the surgical incision. However, this is an expected trade-off for obtaining an adaptive agent that behaves according to a particular surgeon's experience in localized incision rather than general incision of a random location in the patient's cornea.

 \subsubsection{Adaptive Performance}
 Although SCR slightly decreases with incentivized adaptation, the opposite effect can be observed for the adaptation metric (AdSSR). Table~\ref{tab:adssr} shows a clear and significant increase in AdSSR with higher values of $\lambda_{gail}$ compared to $\lambda_{env}$. The agent adapted to the left incision approach shows an increase in AdSSR of 150\% of its original value, while the agent adapted to the right incision approach shows an increase of 16.6\%. This imbalance can be explained by the initially skewed AdSSR of the \textit{NonAdaptAgent} being 28\% and 72\% for left and right, respectively. 
 However, even with a relatively small increase of 16.6\% of the right incision approach, it still surpasses the maximum decrease of the SCR value of 16\%. This indicates that while the approach may impact the agent's proficiency in executing surgical incisions, it significantly enhances the adaptability indicator. Fig.~\ref{fig:scr_vs_adssr} shows this trade-off between adaptability and completion rate for left and right half incision.


\section{Conclusion and Future work}
\label{sec:conclusion}
In conclusion, this work introduces an innovative approach to enhance the precision and adaptability of autonomous surgical agents, particularly in the context of ophthalmic cataract surgery. The proposed method utilizes a simulated environment to teach an agent how to mimic surgeon style and performance through reinforcement, imitation, and curriculum learning techniques.
Additionally, we laid the groundwork for several exciting avenues of future work. The imitation learning view of our work paves the way for online real-time learning, allowing autonomous agents to continuously adapt and refine their skills based on live demonstrations from surgeons. This approach allows agents to stay up-to-date with evolving surgical techniques and adapt to specific situations. 
Concurrently, our curriculum learning-based approach allows a seamless transition from simulation environments to physical robots operating on synthetic porcine eyes. Our work represents a crucial step towards practical implementation in clinical settings. 
Furthermore, our work moves towards creating a single model capable of understanding and isolating various surgical approaches, responding to control input signals to switch between them as needed. This dynamic adaptability would enable autonomous surgical agents to cater to different surgeons' preferences and adapt to various surgical scenarios. 
In summary, this work aimed to address the challenges of adaptability and precision; it paves the way for future advancements to improve patient outcomes, reduce surgical errors, and improve the overall surgical experience.







\section*{ACKNOWLEDGMENT}

This work is partially funded by the German Ministry of Education and Research (BMBF) under the TeachTAM project (Grant Number: 01IS17043).


\bibliographystyle{IEEEtran}
\bibliography{IEEEabrv,IEEEexample}

\end{document}